\documentclass{article}

\usepackage{arxiv}

\usepackage[utf8]{inputenc} 
\usepackage[T1]{fontenc}    
\usepackage{hyperref}       
\usepackage{url}            
\usepackage{booktabs}       
\usepackage{amsfonts}       
\usepackage{nicefrac}       
\usepackage{microtype}      
\usepackage{graphicx}
\usepackage{natbib}
\usepackage{amsmath}
\usepackage{doi}
\usepackage{amsthm}

\title{On the High Symmetry of Neural Network Functions}

\author{ \href{https://orcid.org/0000-0002-6060-5365}{\includegraphics[scale=0.06]{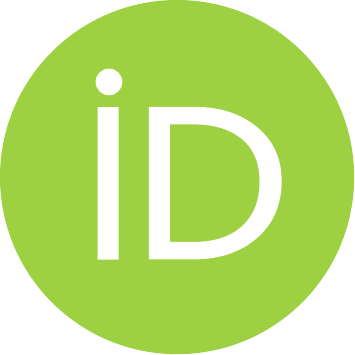}\hspace{1mm}Umberto Michelucci}\thanks{\url{https://www.toelt.ai}} \\
    Research and Development \\
	TOELT LLC\\
	Switzerland \\
	\texttt{umberto.michelucci@toelt.ai} \\[0.3cm]
	Department of Computer Science\\
	Lucerne University of Applied Sciences and Arts\\
	Switzerland }

\begin{document}
\newtheorem{definition}{Definition}[section]
\newtheorem{theorem}{Theorem}[section]
\maketitle

\begin{abstract}
Training neural networks means solving a high-dimensional optimization problem. Normally the goal is to minimize a loss function that depends on what is called the network function, or in other words the function that gives the network output given a certain input. This function depends on a large number of parameters, also known as weights, that depends on the network architecture. In general the goal of this optimization problem is to find the global minimum of the network function. In this paper it is discussed how due to how neural networks  are designed, the neural network function present a very large symmetry in the parameter space. This work shows how the neural network function has a number of equivalent minima, in other words minima that give the same value for the loss function and the same exact output, that grows factorially with the number of neurons in each layer for feed forward neural network or with the number of filters in a convolutional neural networks. When the number of neurons and layers is large, the number of equivalent minima grows extremely fast. This will have of course consequences for the study of how neural networks converges to minima during training. This results is known, but in this paper for the first time a proper mathematical discussion is presented and an estimate of the number of equivalent minima is derived.
\end{abstract}


\section{Introduction}
Note that this short paper is not meant to review existing results or to analyze deeply the consequences of the high symmetry of neural networks. The only goal is to formalise and calculate the number of equivalent points in parameter space for a feed forward neural network. I hope this can be helpful to someone. The resul highlighted in this short paper (notably without references so far), is known and therefore it is no new contribution. But to the best of my knowledge no one has ever analyzed the problem formally and shown from where this high symmetry is coming from in a short and concise form. I hope the mathematics shown here can be helpful in this regard. I claim not to be the first to have noticed this of course, but I think this paper shows for the first time from where the symmetry comes from.

I plan to add references and search for similar contribution soon and update this short paper accordingly.

The contribution of this short paper is the {\sl Symmetry Theorem for Feed Forward Neural Networks}, or in other words that for a FFNN with $L$ layers the network function $\hat y(\theta, \textbf{x})$ has a very high symmetry in $\theta$-space. With $\theta$ we have indicated a vector of all the network parameters (or weights), and with $\textbf{x}$ a generic (possibly multi-dimensional) input observation.
More precisely there are $\Lambda = n_1!n_2!...n_{L-1}!$ ($n_i$ being the number of neurons in layer $i$) sets of parameters $\theta$ that gives the exact same value of the network function for the same input $\textbf{x}$. As a direct consequence, the loss function $L$ can not have just one absolute minimum, but always $\Lambda$ ones, all with the same value of $L$. The paper is organised as follows. In Section \ref{sec:formalism} the formalism and notation are discussed. In Section \ref{sec:conclusions} the conclusions and important further research developments are discussed.

\section{Formalism}
\label{sec:formalism}

In this section the case for feed forward neural networks (FFNN) is discussed and the definition of equivalent minima and their exact number depending on the network architecture is respectively given and calculated.

\begin{figure}
    \centering
    \includegraphics[width=14cm]{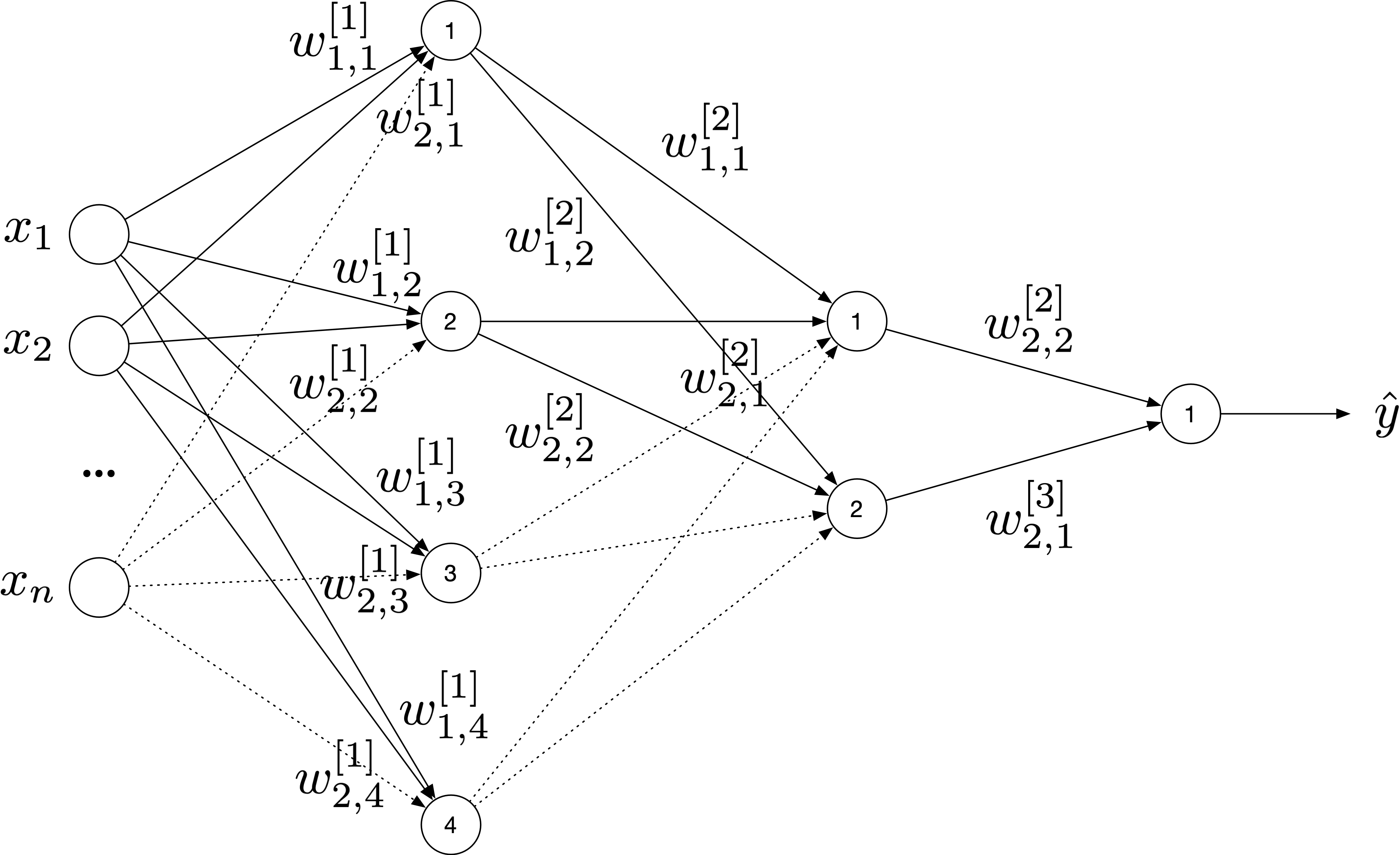}
    \caption{An example of a simple FFNN with two hidden layers. In this diagram only some of the weights (parameters) are depicted for clarity. The symbol $w_{i,j}^{[l]}$ indicates the weight between neuron $i$ in layer $l-1$ and neuron $j$ in layer $l$, where by convention layer 0 is the input layer.}
    \label{fig:ffnn1}
\end{figure}
\begin{figure}[hbt]
    \centering
    \includegraphics[width=14cm]{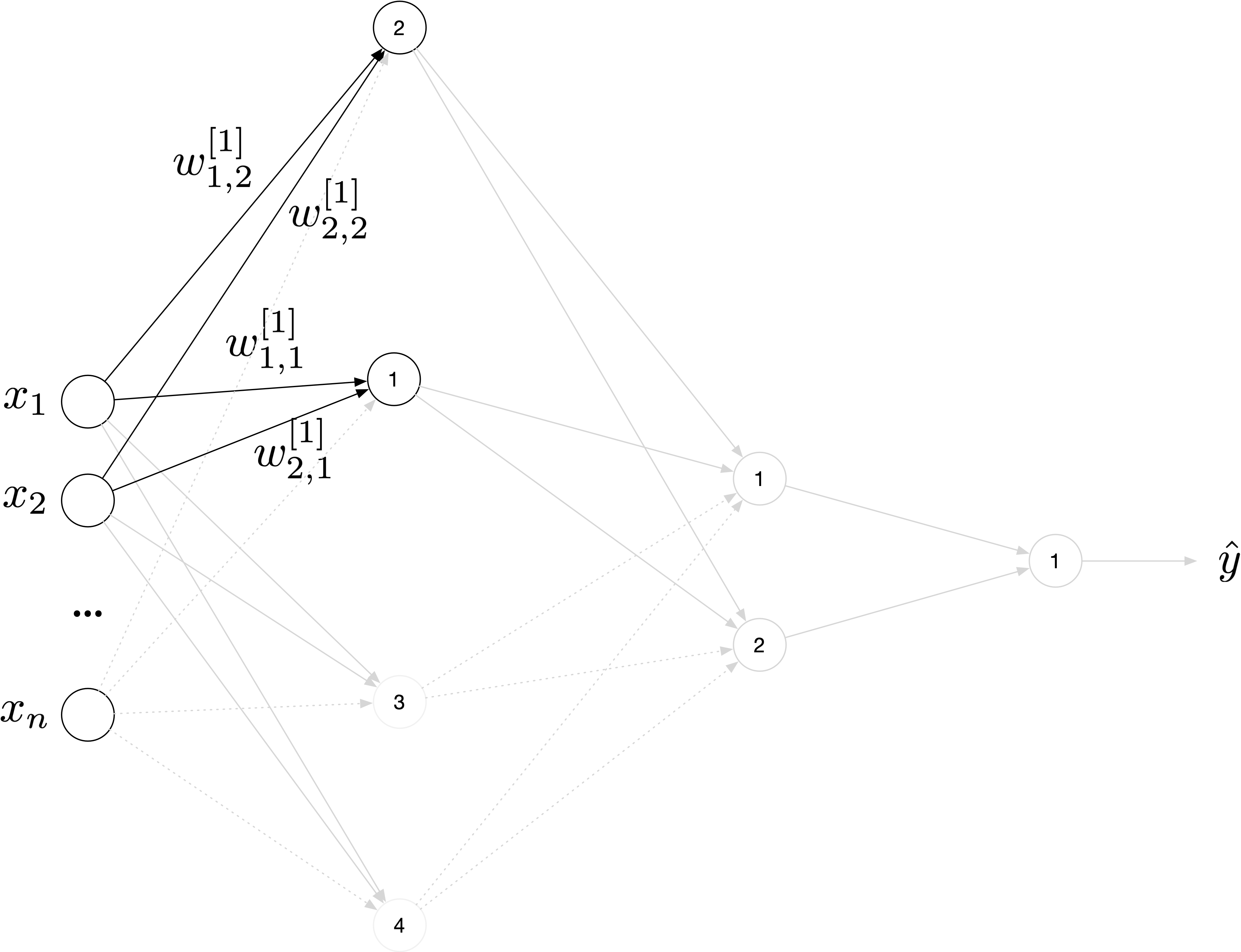}
    \caption{An example of a simple ffnn with two hidden layers. In this diagram neurons 1 and 2 in layer 1 have been switched. The connection with the relative weights have also been switched.}
    \label{fig:ffnn2}
\end{figure}

Let us consider a simple FFNN. An example with just two hidden layers is depicted in Figure \ref{fig:ffnn1}. The symbol $w_{i,j}^{[l]}$ indicates the weight between neuron $i$ in layer $l-1$ and neuron $j$ in layer $l$, where by convention layer 0 is the input layer.
Normally each neuron will also have a bias as input but for simplicity we will neglect them, since the final general results is not influenced by this simplification. We will indicate the output of a generic neuron $i$ in layer $l$ with
\begin{equation}
    a_i^{[l]} = f^{[l]}\left(\sum_{j=1}^{n_{l-1}} w_{j,i}^{[l]} a_j^{[l-1]}\right)
\end{equation}
where $f^{[l]}$ is the activation function in layer $l$. For simplicity we will assume that all layers, except the output one, will have the same activation function $f$. $n_l$ is the number of neurons in layer $l$, $w_{j,i}^{[l]}$ indicates, as already mentioned, the weight between neuron $j$ in layer $l-1$ and neuron $i$ in layer $l$. By convention layer 0 will be the input layer, and therefore $a_j^{[0]}\equiv x_j$, the $j^{th}$ component (or feature) of a generic input observation. In matrix form the output of layer $l$ can be written as
\begin{equation}
   \underbrace{
    \begin{pmatrix}
        a_1^{[l]} \\
        a_2^{[l]} \\
        ... \\
        a_{n_l}^{[l]} 
\end{pmatrix}}_{\displaystyle {\bf a}^{[l]}} =
\underbrace{
    \begin{pmatrix}
        w_{1,1}^{[l]} & w_{2,1}^{[l]} & ... & w_{n_{l-1},1}^{[l]} \\ 
        w_{1,2}^{[l]} & w_{2,2}^{[l]} & ... & w_{n_{l-1},2}^{[l]} \\
        \vdots & ... & \ddots & ... \\
        w_{1,n_l}^{[l]} & w_{2,n_l}^{[l]} & ... & w_{n_{l-1},n_l}^{[l]}
\end{pmatrix}}_{\displaystyle {\bf W}^{[l]}}
\underbrace{\begin{pmatrix}
        a_1^{[l-1]} \\
        a_2^{[l-1]} \\
        ... \\
        a_{n_{l-1}}^{[l-1]} 
\end{pmatrix}}_{\displaystyle {\bf a}^{[l-1]}}
\end{equation}  
or by using the matrix symbols ${\bf a}^{[l]}$ and ${\bf W}^{[l]}$
\begin{equation}
    {\bf a}^{[l]} = {\bf W}^{[l]}{\bf a}^{[l-1]}
\end{equation}
Now let us consider the example in Figure \ref{fig:ffnn1}, and suppose to switch neuron 1 and 2 in layer 1. Suppose to do that by also switching the respective connections, or in other words by switching also the respective weights. To clarify this concept, after switching neurons 1 and 2 in leayer 1, one would end up with the network depicted in Figure \ref{fig:ffnn2} where the irrelevant parts have been colored in light gray to only highlight the changes that the switch have generated.
The neuron switch, performed as described, will have the result of effectively switching the first and second entry in the output vector ${\bf a}^{[1]}$. Note that this can be equivalently achieved by switching rows 1 and 2 in ${\bf W}^{[1]}$. To summarize, switching two generic neurons $i$ and $j$ in layer $l$ as described, is equivalent to switching rows $i$ and $j$ in the weight matrix ${\bf W}^{[l]}$. Switching two neurons is equivalent to a reordering of the neurons in layer $l$, and there are $n_l!$ possible ways of ordering the $n_l$ neurons in layer $l$. Now is important to note that by switching two neurons as described will give naturally a different output of the network function. But we can easily correct this by simply switching columns $i$ and $j$ in ${\bf W}^{[l+1]}$. To see why this is the case let us consider switching neuron $i$ and $j$ in layer $l$ as described. This will result in an output vector ${\bf a'}^{[l]}$ given by
\begin{equation}
    {\bf a'}^{[l]} = \begin{pmatrix}
        a_1^{[l]} \\
        a_j^{[l]} \\
        ... \\
        a_i^{[l]} \\
        ... \\
        a_{n_l}^{[l]} 
\end{pmatrix}
\end{equation}
where rows $i$ and $j$ has been swapped. Now let us consider what happens to the output of layer $l+1$. If ${\bf W}^{[l+1]}$ is not changed the output vector ${\bf a}^{[l+1]}$ will be of course different than the one before the swap of the neurons. But by swapping the columns $i$ and $j$ in ${\bf W}^{[l+1]}$ and indicating the new weight matrix with ${\bf W}_c^{[l]}$, the output vector ${\bf a}^{[l+1]}$ will remain unchanged. In fact
\begin{equation}
   \underbrace{
    \begin{pmatrix}
        a_1^{[l+1]} \\
        a_2^{[l+1]} \\
        ... \\
        a_{n_{l+1}}^{[l+1]} 
\end{pmatrix}}_{\displaystyle {\bf a}^{[l]}} =
\underbrace{\begin{pmatrix}
        w_{1,1}^{[l]} & ... & w_{j,1}^{[l]} & ... & w_{i,1}^{[l]} & ...&w_{n_{l-1},1}^{[l]}\\
        w_{1,2}^{[l]} & ... & w_{j,2}^{[l]} & ... &w_{i,1}^{[l]} & ...&w_{n_{l-1},2}^{[l]}\\
        ... \\
        w_{1,n_l}^{[l]} & ... &  w_{j,n_l}^{[l]} & ... &w_{i,1}^{[l]} &...& w_{n_{l-1},n_l}^{[l]}\\
\end{pmatrix}}_{\displaystyle {\bf W}_c^{[l]}}
    \underbrace{\begin{pmatrix}
        a_1^{[l]} \\
        a_j^{[l]} \\
        ... \\
        a_i^{[l]} \\
        ... \\
        a_{n_l}^{[l]} 
\end{pmatrix}}_{\displaystyle {\bf a'}^{[l-1]}}
\end{equation}  
That means that by simultaneosuly swapping two rows $i$ and $j$ in ${\bf W}^{[l]}$, and indicating the new weight matrix with ${\bf W}_r^{[l]}$, and the two columns $i$ and $j$ in ${\bf W}^{[l+1]}$ the output of the network will not change. We have effectively found two set of weights 
$$\{ {\bf W}^{[0]}, {\bf W}^{[1]}, ..., {\bf W}^{[L]}\}$$ 
and  
$$\{ {\bf W}^{[0]}, {\bf W}^{[1]}, ..., {\bf W}_r^{[l]}, {\bf W}_c^{[l+1]}, ..., {\bf W}^{[L]}\}$$
that will give the exact same value for the loss function and for all the predictions. As discussed there are $n_l!$ possible ways of organising the neurons in layer $l$ that will give a set of weights with the property of keeping the neural network output and loss function unchanged. One can say that the neural network is invariant to neuron swaps as described in the text. This property is general for all layers (excluded the input and the output) $1,...,L-1$. So in total one can create $\Lambda = n_1! n_2! ... n_{L-1}!$ possible set of weights that will give the exact same predictions and loss function value. In a small network with three hidden layers, each with 128 neurons the number of equivalent set of weights is $128!^3$ that is approximately $5.7 \cdot 10^{646}$. This is very large number indicating that in the parameter space the loss function will have this number of {\sl equivalent} minima. Where again with equivalent minima we intend location in parameter space that will give the same value of the loss function and the exact same predictions.

Let us formally define a {\sl generalized neuron switch}.
\begin{definition}
A {\sl generalized $i$-$j$ neuron switch} is a transformation of the set of weights that is described by
\begin{equation}
\left\{
    \begin{matrix}
        {\bf W}^{[l]} \xrightarrow{R_i \rightarrow R_j} {\bf W}_r^{[l]} \\
        {\bf W}^{[l+1]} \xrightarrow{C_i \rightarrow C_j} {\bf W}_c^{[l+1]} \\
    \end{matrix} \right.
\end{equation}

where the notation $\xrightarrow{R_i \rightarrow R_j}$ indicates the switch of rows $i$ and $j$ and $\xrightarrow{C_i \rightarrow C_j}$ the switch of columns $i$ and $j$. This matrix transformation is equivalent to switch neurons $i$ and $j$ in layer $l$ as described in the text.
\end{definition}

One can say that $L$ and $\hat y$ are invariant under a {\sl generalized $i$-$j$ neuron switch}. There are a total of $\Lambda = n_1! n_2! ... n_{L-1}!$ generalized $i$-$j$ neuron switches possible in a FFNN with $L$ layers (including the output layer). 
Let us also define the concept of {\sl equivalent set of parameters} for a given FFNN.
\begin{definition}
\label{def:2}
Two set of parameters $\overline {\bf W}_1 = \{ {\bf W}_1^{[0]}, {\bf W}_1^{[1]}, ..., {\bf W}_1^{[L]}\}$) and $\overline {\bf W}_2 = \{ {\bf W}_2^{[0]}, {\bf W}_2^{[1]}, ..., {\bf W}_2^{[L]}\}$)
are said to be equivalent for a given FFNN when $L(\overline {\bf W}_1) = L(\overline {\bf W}_2)$ and $\hat y(\overline{\bf W}_1)= \hat y(\overline{\bf W}_2)$ for any input ${\bf x} = (x_1, ..., x_n)$.
\end{definition}

So the conclusion of this paper can be stated in the following theorem.
\begin{theorem}[Symmetry Theorem for Feed Forward Neural Networks]
Given a FFNN with $L$ layers, each having $n_i$ neurons, there exist $\Lambda = n_1!n_2!...n_{L-1}!$ equivalent sets of weights.
\end{theorem}
The proof has been given in the previous discussion.

\subsection{Convolutional Neural Neural Networks}
The same approach can be used for convolutional neural networks (CNNs). In fact the order of the filters in each convolution is not relevant and therefore also in CNNs we have a high symmetry. this time the relevant parameters are not the number of neurons in each layer, but the number of filters in each convolutional layer. Details for CNNs will be developed by the author in a subsequent publication.

\section{Conclusions}
\label{sec:conclusions}

For any point in parameter space $\overline {\bf W} = \{ {\bf W}^{[0]}, {\bf W}^{[1]}, ..., {\bf W}^{[L]}\}$ there are always $\Lambda-1$ additional points in parameter space that give the same loss function value and the same network output as $\overline {\bf W}$. This highlight the very high symmetry of the network function in parameter space as the number $\Lambda$ grows as $O(n!^L)$ with the number of neurons $n$ in a given layer and $L$ the total number of layers. That also means that if the network function cannot have one single minima, but if there is one then there will be $\Lambda$ that will be equivalent according to Defintion \ref{def:2}.







\end{document}